\documentclass{article}
\usepackage[utf8]{inputenc}
\usepackage{amsmath}
\usepackage{amssymb}
\usepackage[title]{appendix}
\usepackage{authblk}
\usepackage[]{caption} 
\usepackage{cite}
\usepackage{enumitem}
\usepackage{etoolbox}
\usepackage[hang,flushmargin]{footmisc}
\usepackage{geometry}
\usepackage{graphicx}
\usepackage{hyperref}
\usepackage{longtable}
\usepackage{mathtools}
\usepackage{multicol}
\usepackage{nomencl}
\usepackage{siunitx}
\usepackage{parskip}
\usepackage{subcaption}
\usepackage{tabularx}
\usepackage{url}

\graphicspath{{figs/}}

\title{Signal Enhancement for Magnetic Navigation \\ Challenge Problem}

\author[1]{Albert R. Gnadt}
\author[2]{Joseph Belarge}
\author[3]{Aaron Canciani}
\author[2]{Glenn Carl}
\author[2]{Lauren Conger}
\author[3]{Joseph Curro}
\author[1]{Alan Edelman}
\author[2]{Peter Morales}
\author[3]{Aaron P. Nielsen}
\author[2]{Michael F. O'Keeffe}
\author[1]{Christopher V. Rackauckas}
\author[2]{Jonathan Taylor}
\author[2]{Allan B. Wollaber}

\affil[1]{Massachusetts Institute of Technology, Cambridge, MA, USA}
\affil[2]{MIT Lincoln Laboratory, Lexington, MA, USA}
\affil[3]{Air Force Institute of Technology, Wright-Patterson AFB, OH, USA}

\date{January 6, 2023}

\begin{document}
\maketitle

\section{Introduction}

Harnessing the magnetic field of the Earth for navigation has shown promise as a viable alternative to other navigation systems. Commercial and government organizations have surveyed the Earth to varying degrees of precision by collecting and storing magnetic field data in the form of magnetic anomaly maps. A magnetic navigation system collects its own magnetic field data using a magnetometer and uses magnetic anomaly maps to determine the current location. This technique does not rely on satellites or other external communications, and it is available globally at all times and in all weather.

The greatest challenge with magnetic navigation arises when the magnetic field measurements from the magnetometer encompass the magnetic field from not just the Earth, but also from the vehicle on which it is mounted. The total magnetic field is a linear superposition of the magnetic fields of the vehicle and the Earth with additional contributions from diurnal variation (i.e., space weather), which can be largely removed using ground-based reference measurements. However, magnetometers report the net magnetic field vector. It is difficult to separate the Earth magnetic anomaly field, which is crucial for navigation, from the total magnetic field reading from the sensor.

The purpose of this challenge problem is to decouple the Earth and aircraft magnetic signals in order to derive a clean signal from which to perform magnetic navigation. Baseline testing on the dataset has shown that the Earth magnetic field can be extracted from the total magnetic field using machine learning (ML). The challenge is to remove the aircraft magnetic field from the total magnetic field using a trained model. This challenge offers an opportunity to construct an effective model for removing the aircraft magnetic field from the dataset by using a scientific machine learning approach comprised of an ML algorithm integrated with the physics of magnetic navigation.

\newpage

\section{Magnetic Navigation Background}

Magnetic navigation is enabled by variations in the crustal magnetic field of the Earth, also known as the magnetic anomaly field. The total geomagnetic field is comprised of fields from several sources. The dominant source is the core field with values of $25,000 - 65,000$ nT at the surface of the Earth, about $100\times$ weaker than a refrigerator magnet. The magnetic anomaly field magnitude is typically less than 1000 nT, about $100\times$ weaker than the core field. As such, magnetic navigation requires the ability to sense small differences in the magnetic anomaly field, which can be mapped and is stable over geological time scales. The spatial extent of the crustal magnetic fields make them strong enough for navigation even at altitudes of tens of kilometers above the Earth's surface.

The strength of a static magnetic field arising from a localized source follows the inverse cubic distance scaling law of a magnetic dipole (when the distance to the source is much larger than the spatial extent of the source). This high drop-off rate in magnetic fields means that it is difficult for disturbances to affect magnetic sensors from a distance without exhorting a significant amount of power, making it difficult to interfere with or jam magnetic navigation from ground stations. Thus, the predominant issue with magnetic navigation comes from magnetic interference generated by the aircraft itself. The purpose of this challenge is to effectively remove the magnetic interference of the aircraft from the readings of the on-board magnetometers so that effective magnetic navigation can be performed.

Traditionally the Earth and aircraft magnetic fields can be separated using the Tolles-Lawson model \cite{Tolles1950,Tolles1954,Tolles1955}, as detailed in \cite{Gnadt2022c}. Determining the coefficients of the model hinges on first collecting magnetometer data during a calibration flight involving a set of pitch, roll, and yaw maneuvers performed at a high altitude over a region with a small magnetic gradient. As part of the Tolles-Lawson model, a bandpass filter is applied to these scalar and vector magnetometer measurements and several crucial assumptions about the nature of the aircraft magnetic field are made, such as:

\begin{enumerate}
    \item The magnetic sources on the aircraft arise from permanent dipole, induced dipole, and eddy current fields.
    \item The permanent dipole sources do not change over time.
    \item The induced dipoles depend on the orientation of the aircraft with respect to the magnetic field of the Earth.
    \item The inductance from electrical current paths is zero, so the eddy currents arise from instantaneous changes of the magnetic flux through a surface.
\end{enumerate}

These assumptions are sufficient when the magnetometer is placed on a 3m boom behind the aircraft (tail stinger), because the magnetic field from the aircraft is weak enough relative to the Earth at the sensor. However, this is impractical for operational aircraft. The Tolles-Lawson model does not produce a compensated signal with sufficiently accurate results when the magnetometer is close to the multiple magnetic interference sources of the aircraft, such as in the cabin.

\newpage

\section{The Challenge}

The goal is to take magnetometer readings from within the cabin and remove the aircraft magnetic field to yield a clean magnetic signal. Two options for the desired truth signal are discussed here, though others may be possible. One such truth signal is the tail stinger magnetometer which, after professional compensation, is sufficiently accurate for magnetic navigation. It has much less corruption than the in-cabin sensors due to its location far aft of the cockpit, control surfaces, and other sources of significant magnetic interference. However, an aircraft with a tail stinger must be available for this option. The other option is to treat the magnetic anomaly map in the collection region as the truth signal. Using the path of the aircraft, the magnetic anomaly signal over this path can be determined and treated as the truth signal. However, an accurate magnetic anomaly map must be available for this option.

The first case wherein the tail stinger is treated as the truth signal has several advantages. The primary advantage is that there is no need to potentially account for for minor position inaccuracies. Additionally, no interpolations are required to determined the truth signal, which may lead to errors. Finally, most conditions, such as weather, that were present during the collection would be identical for all sensors meaning that additional compensation due to known conditions would likely be unnecessary. Thus, this option is used for this challenge problem.

\section{Judging Criteria}

For this challenge problem, an evaluation (testing) dataset is withheld from participants. The goal is to learn a model for removing the aircraft magnetic field from the build (training) dataset that generalizes to the evaluation dataset. \textbf{Submissions are judged based on the standard deviation of the error between the model output and the truth signal on a per-flight line basis.} The truth signal is \texttt{mag\_1\_c}, the professionally compensated tail stinger scalar magnetometer measurement. This challenge problem was introduced at JuliaCon 2020 and will remain open through 2023. All submissions should be sent to \href{mailto:magnav-admins@mit.edu}{magnav-admins@mit.edu} in the form of a Git repository with code that can perform predictions on the evaluation dataset without any modifications to the source code required. Submissions must contain a script that accepts any of the original HDF5 files as input and produces flight line, time, and compensated data, in tabular format, as output. It is recommended to include a short (e.g., 5-page) report that explains the methodology.

\section{Flight Data Description}

A campaign was conducted during the summer of 2020 to collect magnetic field data. During the data collection, flight patterns were planned such that data could be collected at various altitudes. This is typically measured in height above ellipsoid (HAE), where the aircraft is flying at a constant altitude above the Earth's assumed perfectly ellipsoidal shape. For drape surfaces, the aircraft is flying at an approximately constant altitude above ground level (AGL), i.e. over actual crustal features, such as mountains and valleys.

\newpage

\begin{figure}[ht]
    \centering
    \includegraphics[width=0.90\textwidth]{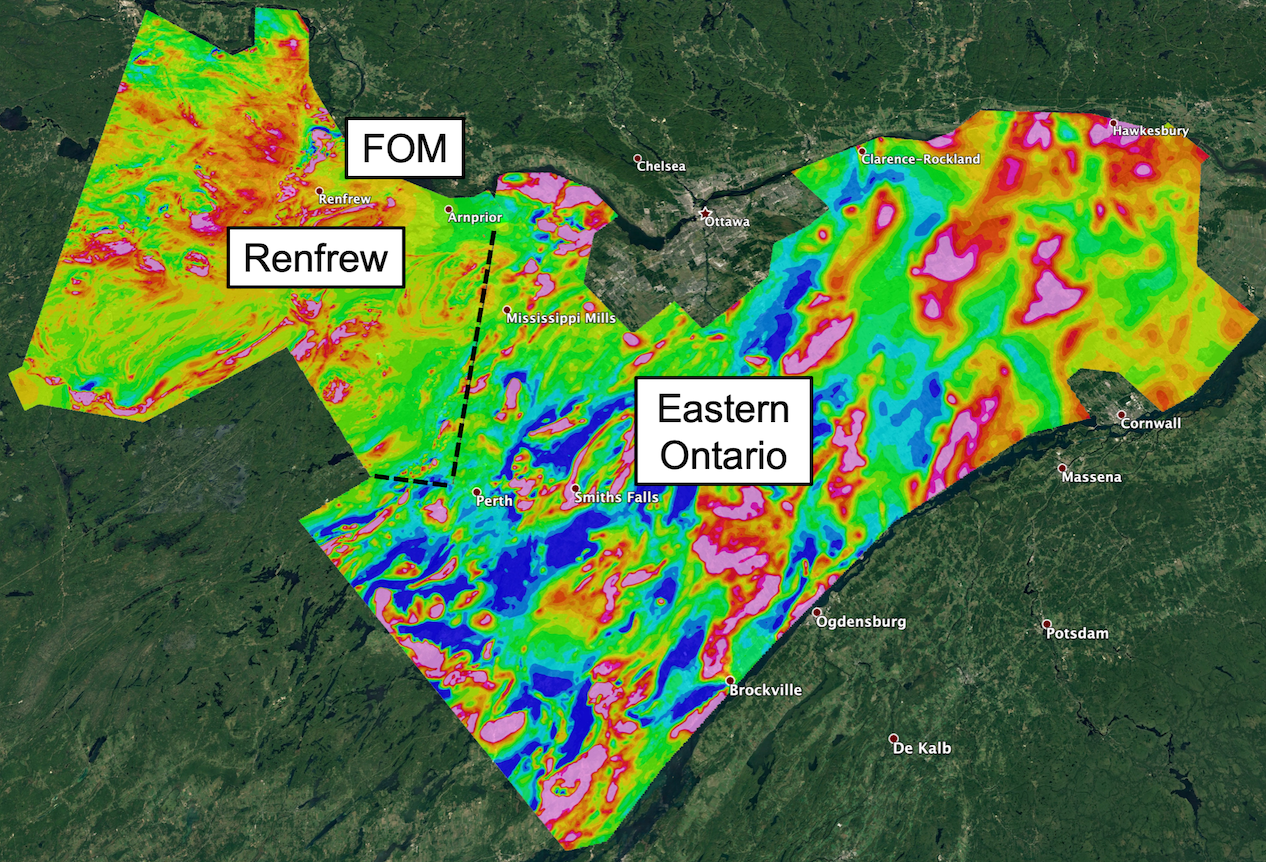}
    \caption{Magnetic anomaly maps near Ottawa, Ontario, Canada. The western region is the Renfrew flight area. The eastern region is the Eastern Ontario flight area. The northern region is the figure of merit (FOM) flight area.}
    \label{fig:sgl_map}
\end{figure}

The measurements were collected by Sander Geophysics Ltd.\ (SGL) \cite{SGL2020} near Ottawa, Ontario, Canada, using a Cessna Grand Caravan equipped with a number of sensors. The measurements were collected in three flight areas, which are shown in Figure~\ref{fig:sgl_map}. Scalar measurements of the total field were generated from five optically pumped, split-beam cesium vapor magnetometers. Four fluxgate magnetometers, one at the base of the tail stinger and three inside the aircraft, were also used for vector measurements of the total field. One scalar magnetometer was positioned on a tail stinger to collect magnetic measurements with minimal aircraft magnetic field noise. The remaining four scalar magnetometers were placed inside the cabin of the aircraft. The locations of the sensors within the cabin can be found in Table \ref{tab:sensor_loc}.

\begin{table}[ht]
    \centering
    \caption[Summary of scalar and vector magnetometer locations]{Summary of scalar and vector magnetometer locations. The reference point is the front seat rail. $x$ is positive in the aircraft forward direction, $y$ is positive to port (left facing forward), and $z$ is positive upward. Note that each vector magnetometer is arbitrarily oriented.}
    \label{tab:sensor_loc}
    \begin{tabular}{c l c c c}
        \hline
        Sensor Name & Location                      &  $x$ [m]  & $y$ [m] & $z$ [m] \\ \hline
        & \multicolumn{2}{c}{Scalar Magnetometers} \\ \hline
        Mag 1       & Tail stinger                  & -12.01  &  0    & 1.37 \\
        Mag 2       & Front cabin, aft of cockpit   &  -0.60  & -0.36 & 0 \\
        Mag 3       & Mid cabin, near INS           &  -1.28  & -0.36 & 0 \\
        Mag 4       & Rear cabin, floor             &  -3.53  &  0    & 0 \\
        Mag 5       & Rear cabin, ceiling           &  -3.79  &  0    & 1.20 \\ \hline
        & \multicolumn{2}{c}{Vector Magnetometers} \\ \hline
        Flux A      & Mid cabin, near fuel tank     &  -3.27  & -0.60 & 0 \\
        Flux B      & Tail, base of stinger         &  -8.92  &  0    & 0.96 \\ 
        Flux C      & Rear cabin, port              &  -4.06  &  0.42 & 0 \\
        Flux D      & Rear cabin, starboard         &  -4.06  & -0.42 & 0 \\
    \end{tabular}
\end{table}

In addition to the magnetometers, supplemental sensors collected relevant flight data. A subset of this data, such as compensated tail stinger measurements, contain information redundant to the truth signal. Using this data could lead to falsely accurate models, and therefore should not be used. The data fields shown below were determined to provide information for training, while not providing direct truth data. For information on all of the data fields collected, see Appendix \ref{sec:data_fields}. Note that Earth's core field from the International Geomagnetic Reference Field (IGRF) \cite{Alken2021} can be calculated along the flight path as \texttt{igrf = mag\_1\_dc $-$ mag\_1\_igrf}.

\newpage

\begin{multicols}{3}
\begin{itemize}
    \item diurnal
    \item igrf
    \item mag\_2\_uc
    \item mag\_3\_uc
    \item mag\_4\_uc
    \item mag\_5\_uc
    \item flux\_a\_x
    \item flux\_a\_y
    \item flux\_a\_z
    \item gradient(flux\_a\_x)
    \item gradient(flux\_a\_y)
    \item gradient(flux\_a\_z)
    \item flux\_b\_x
    \item flux\_b\_y
    \item flux\_b\_z
    \item gradient(flux\_b\_x)
    \item gradient(flux\_b\_y)
    \item gradient(flux\_b\_z)
    \item flux\_c\_x
    \item flux\_c\_y
    \item flux\_c\_z
    \item gradient(flux\_c\_x)
    \item gradient(flux\_c\_y)
    \item gradient(flux\_c\_z)
    \item flux\_d\_x
    \item flux\_d\_y
    \item flux\_d\_z
    \item gradient(flux\_d\_x)
    \item gradient(flux\_d\_y)
    \item gradient(flux\_d\_z)
    \item ins\_acc\_x
    \item ins\_acc\_y
    \item ins\_acc\_z
    \item ins\_pitch
    \item ins\_roll
    \item ins\_yaw
    \item gradient(ins\_pitch)
    \item gradient(ins\_roll)
    \item gradient(ins\_yaw)
    \item cur\_com\_1
    \item cur\_ac\_hi
    \item cur\_ac\_lo
    \item cur\_tank
    \item cur\_flap
    \item cur\_strb
    \item cur\_srvo\_o
    \item cur\_srvo\_m
    \item cur\_srvo\_i
    \item cur\_heat
    \item cur\_acpwr
    \item cur\_outpwr
    \item cur\_bat\_1
    \item cur\_bat\_2
    \item vol\_acpwr
    \item vol\_outpwr
    \item vol\_bat\_1
    \item vol\_bat\_2
    \item vol\_res\_p
    \item vol\_res\_n
    \item vol\_back\_p
    \item vol\_back\_n
    \item vol\_gyro\_1
    \item vol\_gyro\_2
    \item vol\_acc\_p
    \item vol\_acc\_n
    \item vol\_block
    \item vol\_back
    \item vol\_servo
    \item vol\_cabt
    \item vol\_fan
\end{itemize}
\end{multicols}

\smallskip

Data from six flights is publicly available. Each flight contained a different set of objectives, and as such the data from each flight has individual nuances. The details of Flight 1003 are discussed below. For further information on Flights 1002, 1004, 1005, 1006, and 1007, see Appendix \ref{sec:other_flights} and the \texttt{readmes} folder in the MagNav.jl \cite{Gnadt2022a} repository.

The objective for Flight 1003 (see \texttt{Flt1003\_readme.txt}) was to measure the crustal magnetic field at two altitudes (400m and 800m) in both the Eastern Ontario and Renfrew flight areas. The flight was conducted on June 29, 2020 and lasted approximately 5 hr and 42 min. A summary of the flight lines (segments) is shown in Table \ref{tab:flight_1003}, Here it can be seen that flight line 1003.10 has been withheld for the evaluation dataset, while the remaining lines are provided in the build dataset.

\begin{table}[ht]
  \caption{Flight lines for Flight 1003.}
  \label{tab:flight_1003}
  \begin{center}
    \begin{tabular}{l l}
    \hline
    Line & Description \\ \hline
    1003.01     & Takeoff - Eastern Ontario Free-Fly \\
    1003.02     & Eastern Ontario Free-Fly 400m      \\
    1003.03     & Climb to 800m                      \\
    1003.04     & Eastern Ontario Free-Fly 800m      \\
    1003.05     & Transit at 800m                    \\
    1003.06     & Descend to 400m                    \\
    1003.07     & Transit to Renfrew Free-Fly        \\
    1003.08     & Renfrew Free-Fly 400m              \\
    1003.09     & Climb to 800m                      \\
    1003.10     & HOLD-OUT TESTING DATA              \\
    1003.11     & Transit to base                    \\
    \hline
    \end{tabular}
  \end{center}
\end{table}

\newpage

\section{MagNav.jl Overview}

MagNav.jl \cite{Gnadt2022a} is an open-source software package for aeromagnetic compensation and airborne magnetic anomaly navigation written in the Julia programming language \cite{Bezanson2017}. It was developed almost entirely by the lead author, but largely based on work by Aaron Canciani at the Air Force Institute of Technology \cite{Canciani2016a}. The package functionalities can be divided into the four essential components required for MagNav: sensors (flight data), magnetic anomaly maps, aeromagnetic compensation models, and navigation algorithms. The package is designed to be robust for different flight datasets, though it is based on the dataset described in this work.

\begin{figure}[ht]
    \centering
    \includegraphics[width=0.90\textwidth]{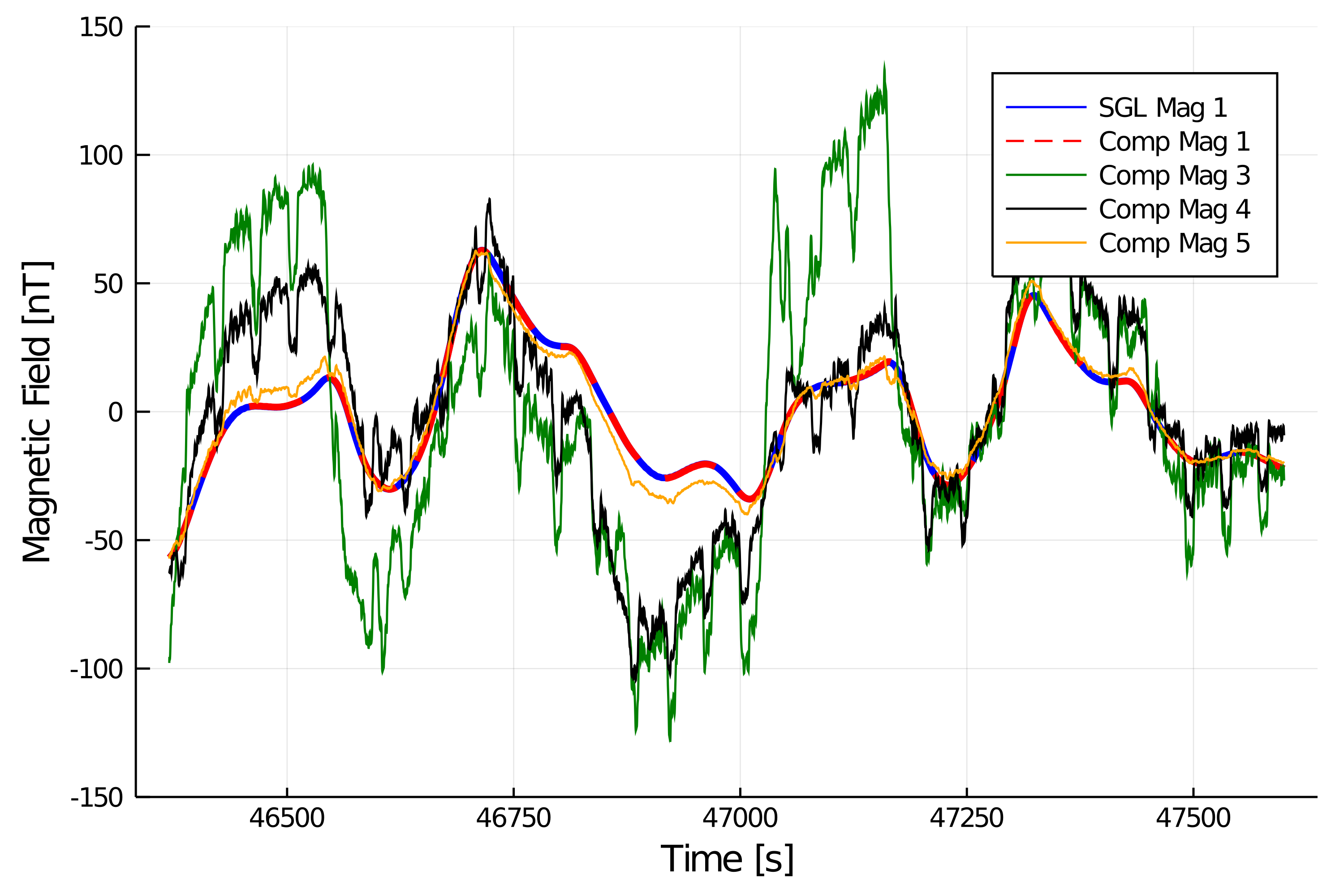}
    \caption{Flight 1002 magnetometers compensated using the Tolles-Lawson model.}
    \label{fig:comp_prof_1}
\end{figure}

Sample run files have also been provided in the MagNav.jl \cite{Gnadt2022a} repository within the \texttt{runs} folder. After running the \texttt{example\_sgl.jl} file, Figure~\ref{fig:comp_prof_1} should be plotted as a baseline result. Here, the scalar magnetometers have been compensated using the Tolles-Lawson model (function \texttt{create\_TL\_coef}), then detrended (function \texttt{detrend}) to remove sensor biases. As shown in Figure~\ref{fig:comp_prof_1}, the magnetometer 1 (tail stinger) compensation exactly matches the professional compensation done by SGL (truth signal). Magnetometer 5, despite being located in the cabin, also performs fairly well, while magnetometers 3 and 4 would not be suitable for magnetic navigation. Magnetometer 2 is not shown due to even worse performance.

\newpage

\begin{appendices}

\section{SGL 2020 Flight Data Fields} \label{sec:data_fields}

In addition to scalar and vector magnetometer measurements, various auxiliary sensor data from the SGL flights is included in the datasets. Below is a description of each available data field. (WGS-84) indicates elevation above the WGS-84 ellipsoid.

\begin{longtable}{|r|r|l|}
    \caption[SGL 2020 flight data fields]{SGL 2020 flight data fields.}
    \label{tab:data2020}
    \\ \hline
    \textbf{Field} & \textbf{Units} & \textbf{Description} \\ \hline
    tie\_line & - & line number \\
    flight & - & flight number \\
    year & - & year \\
    doy & - & day of year \\
    tt & s & fiducial seconds past midnight UTC \\
    utmX & m & x-coordinate, WGS-84 UTM zone 18N \\
    utmY & m & y-coordinate, WGS-84 UTM zone 18N \\
    utmZ & m & z-coordinate, GPS altitude (WGS-84) \\
    msl & m & z-coordinate, GPS altitude above EGM2008 Geoid \\
    lat & deg & latitude, WGS-84 \\
    lon & deg & longitude, WGS-84 \\
    baro & m & barometric altimeter \\
    radar & m & filtered radar altimeter \\
    topo & m & radar topography (WGS-84) \\
    dem & m & digital elevation model from SRTM (WGS-84) \\
    drape & m & planned survey drape (WGS-84) \\
    ins\_pitch & deg & INS-computed aircraft pitch \\
    ins\_roll & deg & INS-computed aircraft roll \\
    ins\_yaw & deg & INS-computed aircraft yaw \\
    diurnal & nT & measured diurnal \\
    mag\_1\_c & nT & Mag 1: compensated magnetic field \\
    mag\_1\_lag & nT & Mag 1: lag-corrected magnetic field \\
    mag\_1\_dc & nT & Mag 1: diurnal-corrected magnetic field \\
    mag\_1\_igrf & nT & Mag 1: IGRF \& diurnal-corrected magnetic field  \\
    mag\_1\_uc & nT & Mag 1: uncompensated magnetic field \\
    mag\_2\_uc & nT & Mag 2: uncompensated magnetic field \\
    mag\_3\_uc & nT & Mag 3: uncompensated magnetic field \\
    mag\_4\_uc & nT & Mag 4: uncompensated magnetic field \\
    mag\_5\_uc & nT & Mag 5: uncompensated magnetic field \\
    mag\_6\_uc & nT & Mag 6: uncompensated magnetic field \\
    flux\_a\_x & nT & Flux A: fluxgate x-axis \\
    flux\_a\_y & nT & Flux A: fluxgate y-axis \\
    flux\_a\_z & nT & Flux A: fluxgate z-axis \\
    flux\_a\_t & nT & Flux A: fluxgate total \\
    flux\_b\_x & nT & Flux B: fluxgate x-axis \\
    flux\_b\_y & nT & Flux B: fluxgate y-axis \\
    flux\_b\_z & nT & Flux B: fluxgate z-axis \\
    flux\_b\_t & nT & Flux B: fluxgate total \\
    flux\_c\_x & nT & Flux C: fluxgate x-axis \\
    flux\_c\_y & nT & Flux C: fluxgate y-axis \\
    flux\_c\_z & nT & Flux C: fluxgate z-axis \\
    flux\_c\_t & nT & Flux C: fluxgate total \\
    flux\_d\_x & nT & Flux D: fluxgate x-axis \\
    flux\_d\_y & nT & Flux D: fluxgate y-axis \\
    flux\_d\_z & nT & Flux D: fluxgate z-axis \\
    flux\_d\_t & nT & Flux D: fluxgate total \\
    ogs\_mag & nT & OGS survey diurnal-corrected, levelled, magnetic field \\
    ogs\_alt & m & OGS survey, GPS altitude (WGS-84) \\
    ins\_acc\_x & m/s\textsuperscript{2} & INS x-acceleration \\
    ins\_acc\_y & m/s\textsuperscript{2} & INS y-acceleration \\
    ins\_acc\_z & m/s\textsuperscript{2} & INS z-acceleration \\
    ins\_wander & rad & INS-computed wander angle (ccw from north) \\
    ins\_lat & rad & INS-computed latitude \\
    ins\_lon & rad & INS-computed longitude \\
    ins\_alt & m & INS-computed altitude (WGS-84) \\
    ins\_vn & m/s & INS-computed north velocity \\
    ins\_vw & m/s & INS-computed west velocity \\
    ins\_vu & m/s & INS-computed vertical (up) velocity \\
    pitch\_rt & deg/s & avionics-computed pitch rate \\
    roll\_rt & deg/s & avionics-computed roll rate \\
    yaw\_rt & deg/s & avionics-computed yaw rate \\
    lon\_acc & g & avionics-computed longitudinal (forward) acceleration \\
    lat\_acc & g & avionics-computed lateral (starboard) acceleration \\
    alt\_acc & g & avionics-computed normal (vertical) acceleration  \\
    true\_as & m/s & avionics-computed true airspeed \\
    pitot\_p & kPa & avionics-computed pitot pressure  \\
    static\_p & kPa & avionics-computed static pressure \\
    total\_p & kPa & avionics-computed total pressure \\
    cur\_com\_1 & A & current sensor: aircraft radio 1 \\
    cur\_ac\_hi & A & current sensor: air conditioner fan high \\
    cur\_ac\_lo & A & current sensor: air conditioner fan low \\
    cur\_tank & A & current sensor: cabin fuel pump \\
    cur\_flap & A & current sensor: flap motor \\
    cur\_strb & A & current sensor: strobe lights \\
    cur\_srvo\_o & A & current sensor: INS outer servo \\
    cur\_srvo\_m & A & current sensor: INS middle servo \\
    cur\_srvo\_i & A & current sensor: INS inner servo \\
    cur\_heat & A & current sensor: INS heater \\
    cur\_acpwr & A & current sensor: aircraft power \\
    cur\_outpwr & A & current sensor: system output power \\
    cur\_bat\_1 & A & current sensor: battery 1 \\
    cur\_bat\_2 & A & current sensor: battery 2 \\
    vol\_acpwr & V & voltage sensor: aircraft power \\
    vol\_outpwr & V & voltage sensor: system output power  \\
    vol\_bat\_1 & V & voltage sensor: battery 1 \\
    vol\_bat\_2 & V & voltage sensor: battery 2 \\
    vol\_res\_p & V & voltage sensor: resolver board (+)  \\
    vol\_res\_n & V & voltage sensor: resolver board (-) \\
    vol\_back\_p & V & voltage sensor: backplane (+) \\
    vol\_back\_n & V & voltage sensor: backplane (-) \\
    vol\_gyro\_1 & V & voltage sensor: gyroscope 1 \\
    vol\_gyro\_2 & V & voltage sensor: gyroscope 2 \\
    vol\_acc\_p & V & voltage sensor: INS accelerometers (+)  \\
    vol\_acc\_n & V & voltage sensor: INS accelerometers (-) \\
    vol\_block & V & voltage sensor: block \\
    vol\_back & V & voltage sensor: backplane \\
    vol\_servo & V & voltage sensor: servos \\
    vol\_cabt & V & voltage sensor: cabinet \\
    vol\_fan & V & voltage sensor: air conditioner fan \\ \hline
\end{longtable}

\section{Additional Flight Descriptions} \label{sec:other_flights}

For detailed information on each flight, see the \texttt{readmes} folder in the MagNav.jl \cite{Gnadt2022a} repository. Note that line numbers that do not begin with the flight number refer to repeated survey flight lines. Additionally note that pilot comments were recorded for each flight and are provided in \texttt{pilot\_comments.csv}. This includes time stamps for in-flight events, such as ``FUEL PUMP ON'' and ``POWER LINES,'' which may help explain noise in the magnetic measurements.

The objectives for Flight 1002 (see \texttt{Flt1002\_readme.txt}) were to fly multiple calibration patterns and repeated survey lines, as well as free-fly within mapped regions. The flight began with a high altitude calibration maneuvers in the figure of merit (FOM) flight area over Shawville, Quebec. Then the aircraft flew 3 traverse lines and 2 control lines, in each of the Eastern Ontario and Renfrew flight areas, that had been flown in previous geomagnetic surveys. The purpose of this was to compare the repeat traverse and control lines to the original map data and look for measurement agreement. Additionally, this flight was used to determine that, since there was a trivial difference in accuracy between the traverse and control lines, the map is fully sampled. Starting near the south end of the Eastern Ontario traverse line, the aircraft completed a free-fly portion. This allowed for analysis regarding the amount that upward continuation of drape surface to constant altitude degrades the measurement quality. For this flight, the pilots purposely caused different magnetic events that were expected to alter the virtual dipole of the aircraft. Finally, the aircraft continued to free-fly while changing altitudes, conducted another set of calibration maneuvers, and completed the flight.

\newpage

The objective for Flights 1004 and 1005 (see \texttt{Flt1004\_readme.txt} and \texttt{Flt1005\_readme.txt}) was to create a small, high resolution, recently-flown magnetic anomaly map (the Perth mini-survey) for future comparison and navigation work. A small survey was flown in the Eastern Ontario flight area at 800m. The line spacing was less than the height above ground level (AGL) to ensure that the generated map is fully sampled.

The objective for Flight 1006 (see \texttt{Flt1006\_readme.txt}) was to fly a variety of calibration maneuvers at various altitudes. This included calibration patterns in the FOM flight area at progressively higher altitudes, as well as a low altitude calibration pattern in the Eastern Ontario flight area at 400m. Throughout the flight, the magnetic state of the aircraft was varied.

The objective for Flight 1007 (see \texttt{Flt1006\_readme.txt}) was to free-fly in the Perth mini-survey at 800m and the Eastern Ontario and Renfrew flight areas at 400m.

\end{appendices}
\bigskip
\bigskip
\bigskip

\section*{Acknowledgements}

Research was sponsored by the United States Air Force Research Laboratory and the United States Air Force Artificial Intelligence Accelerator and was accomplished under Cooperative Agreement Number FA8750-19-2-1000. The views and conclusions contained in this document are those of the authors and should not be interpreted as representing the official policies, either expressed or implied, of the United States Air Force or the U.S. Government. The U.S. Government is authorized to reproduce and distribute reprints for Government purposes notwithstanding any copyright notation herein.

The authors appreciate support from the DAF-MIT Artificial Intelligence Accelerator, a joint collaboration between the United States Air Force, MIT CSAIL, and MIT Lincoln Laboratory.

\newpage

\bibliography{main}
\bibliographystyle{IEEEtran}

\end{document}